\RequirePackage{amsmath}
\documentclass[runningheads]{llncs}
\usepackage{graphicx}
\usepackage{layout}
\usepackage{bbold}
\usepackage{lscape}
\usepackage[utf8]{inputenc} 
\usepackage[T1]{fontenc}    
\usepackage{hyperref}       
\usepackage{url}            
\usepackage{booktabs}       
\usepackage{amsfonts}       
\usepackage{amsmath}
\usepackage{nicefrac}       
\usepackage{microtype}      
\usepackage{lipsum}
\usepackage{xcolor}
\usepackage{csquotes}
\usepackage{comment}
\usepackage{graphicx}       
\graphicspath{{media/}}     

\usepackage{listings}
\usepackage{siunitx}
\usepackage{multirow}
\usepackage{cite}
\usepackage{svg}
\usepackage{threeparttable, tablefootnote}
\usepackage{tabularx}

\definecolor{darkred}{rgb}{0.6,0.0,0.0}
\definecolor{darkgreen}{rgb}{0,0.50,0}
\definecolor{lightblue}{rgb}{0.0,0.42,0.91}
\definecolor{orange}{rgb}{0.99,0.48,0.13}
\definecolor{grass}{rgb}{0.18,0.80,0.18}
\definecolor{pink}{rgb}{0.97,0.15,0.45}

\lstdefinestyle{colored}{ %
  basicstyle=\ttfamily,
  backgroundcolor=\color{white},
  commentstyle=\color{green}\itshape,
  keywordstyle=\color{blue}\bfseries\itshape,
  stringstyle=\color{red},
}
\lstdefinelanguage{PythonPlus}[]{Python}{
  morekeywords=[1]{,as,assert,nonlocal,with,yield,self,True,False,None,} 
  morekeywords=[2]{,__init__,__add__,__mul__,__div__,__sub__,__call__,__getitem__,__setitem__,__eq__,__ne__,__nonzero__,__rmul__,__radd__,__repr__,__str__,__get__,__truediv__,__pow__,__name__,__future__,__all__,}, 
  morekeywords=[3]{,object,type,isinstance,copy,deepcopy,zip,enumerate,reversed,list,set,len,dict,tuple,range,xrange,append,execfile,real,imag,reduce,str,repr,}, 
  morekeywords=[4]{,Exception,NameError,IndexError,SyntaxError,TypeError,ValueError,OverflowError,ZeroDivisionError,}, 
  morekeywords=[5]{,ode,fsolve,sqrt,exp,sin,cos,arctan,arctan2,arccos,pi, array,norm,solve,dot,arange,isscalar,max,sum,flatten,shape,reshape,find,any,all,abs,plot,linspace,legend,quad,polyval,polyfit,hstack,concatenate,vstack,column_stack,empty,zeros,ones,rand,vander,grid,pcolor,eig,eigs,eigvals,svd,qr,tan,det,logspace,roll,min,mean,cumsum,cumprod,diff,vectorize,lstsq,cla,eye,xlabel,ylabel,squeeze,}, 
}
\lstdefinelanguage{PyBrIM}[]{PythonPlus}{
  emph={d,E,a,Fc28,Fy,Fu,D,des,supplier,Material,Rectangle,PyElmt},
}

\lstdefinestyle{colorEX}{
  basicstyle=\ttfamily,
  backgroundcolor=\color{white},
  commentstyle=\color{darkgreen}\slshape,
  keywordstyle=\color{blue}\bfseries\itshape,
  keywordstyle=[2]\color{blue}\bfseries,
  keywordstyle=[3]\color{grass},
  keywordstyle=[4]\color{red},
  keywordstyle=[5]\color{orange},
  stringstyle=\color{darkred},
  emphstyle=\color{pink}\underbar,
  captionpos=t
}

\makeatletter
\newcommand{\mypm}{\mathbin{\mathpalette\@mypm\relax}}
\newcommand{\@mypm}[2]{\ooalign{%
  \raisebox{.1\height}{$#1+$}\cr
  \smash{\raisebox{-.6\height}{$#1-$}}\cr}}
\makeatother

\begin{document}
\title{BEExAI: Benchmark to Evaluate Explainable AI}


\author{Samuel Sithakoul\inst{1,2} \and
Sara Meftah\inst{1}\orcidID{0009-0009-3297-5682} \and
Clément Feutry\inst{1}\orcidID{0009-0009-6086-762X}}

\authorrunning{S. Sithakoul et al.}

\institute{
Groupe Square Management, Square Research Center, 173 Avenue Achille Peretti, 92200 Neuilly-sur-Seine, France \and
CentraleSupélec, 3 Rue Joliot Curie, 91190 Gif-sur-Yvette}

\maketitle
	
\setcounter{footnote}{0}
\begin{abstract}
Recent research in explainability has given rise to numerous post-hoc attribution methods aimed at enhancing our comprehension of the outputs of black-box machine learning models.\footnote{There is little consensus regarding the differences between \enquote{interpretability} and \enquote{explainability} \cite{lipton2018mythos}. Within the scope of this paper, we adopt the term \enquote{explainability} to denote methods specifically designed to provide interpretable justifications for the predictions made by a machine learning model.} However, evaluating the quality of explanations lacks a cohesive approach and a consensus on the methodology for deriving quantitative metrics that gauge the efficacy of explainability post-hoc attribution methods. Furthermore, with the development of increasingly complex deep learning models for diverse data applications, the need for a reliable way of measuring the quality and correctness of explanations is becoming critical. We address this by proposing BEExAI, a benchmark tool that allows large-scale comparison of different post-hoc XAI methods, employing a set of selected evaluation metrics. The developed Python library is open-source at \url{https://github.com/SquareResearchCenter-AI/BEExAI}.

 \and Post-hoc methods

\end{abstract}
\section{Introduction}\label{sec:introduction}

Advanced Machine Learning (ML) models, particularly Neural Networks (NNs), are increasingly adopted in a variety of sectors. This is because they outperform simpler models, such as linear regression or ruled-based models, in complex tasks. However, these models are harder to understand, making them less used in high-risk and heavily regulated fields like finance, cybersecurity and healthcare. Hence, there is a growing demand for eXplainable AI methods (XAI methods) to enhance these black-box models’ transparency. In addition, explainability could play an important role in compliance with the AI Act \cite{panigutti2023role}.

Over the past years, different methods have been developed to explain the predictions made by ML models, whose decision making process is not directly explainable. One category, known as ad-hoc methods, aims to train models to generate explanations in parallel with their predictions \cite{lei-etal-2016-rationalizing}. In contrast, post-hoc methods form another family aim to explain models that have already been trained. Unlike ad-hoc methods, post-hoc explanations do not  impact the predictive performance and are, in most cases, model-agnostic. This work particularly focuses on local post-hoc XAI methods that generate attribution scores. Examples of these methods include LIME \cite{ribeiro2016why}, SHAP \cite{lundberg2017unified}, and gradient-based approaches, such as Integrated Gradients \cite{sundararajan2017axiomatic}. Local methods provide explanations for each individual prediction, differing from global methods that offer a broader understanding of the model’s overall behavior. It is worth noting that within the category of local post-hoc methods, there are variations that produce different types of explanations beyond attribution scores. For instance, methods like Anchors \cite{ribeiro2018anchors} extract a sufficient set of rules to generate predictions and methods like Influence functions \cite{koh2017understanding} select training examples that have highly contributed to the generation of a certain prediction of the model.

Nevertheless, even with considerable research efforts in the field, a shared consensus on the quantification and evaluation of the effectiveness of explainability methods is yet to be established \cite{jacovi2020towards}. Indeed, the evaluation stage is often undervalued when working with XAI methods. Users often assume that explanations are sufficient without conducting thorough testing.

In fact, evaluating explanations poses a significant challenge. In various ML areas, benchmarks are frequently proposed and used by the research community \cite{deng2009imagenet,chen2021evaluating,wang2018glue}. These benchmarks propose well-defined metrics and annotated datasets to compare methods and track diverse models and state-of-the-art results, leading to faster progress in research. However, constructing such benchmarks to compare and evaluate XAI methods is a challenging task. The intricacy arises from the difficulty in acquiring reliable ground-truth data. Indeed, when working with \textit{black-box} models, it is difficult for human annotators to predict a model’s inner functioning and the potential connections it may make between inputs and outputs. Therefore, automatic metrics that compare models based on their alignment with human annotations are hard to establish.

In the early stages of explainability research, practitioners often relied on qualitative evaluation to assess how convincing the interpretation is to humans, also known as plausibility. However, this method was insufficient for evaluating the accuracy of the explanation against the actual reasoning process of the model. These evaluations were based on subjective opinions and could vary significantly from one human evaluator to another. Hence, they are inadequate for establishing a reliable reference. 

Recognizing the importance of standardized frameworks for evaluating explanations to ensure the comparability across distinct XAI methods \cite{longo2024explainable}, the scientific community has delved into quantitative evaluation approaches \cite{vilone2021notions,nauta_anecdotal_2023}. As a result, several suggestions for metrics have emerged, structured according to some properties: the \textit{desiderata} of XAI. The main properties that emerge from research works are \textit{Faithfulness} \cite{bhatt_evaluating_2020}, \textit{Plausibility} \cite{zhong2019finegrained}, \textit{Robustness} \cite{alvarez2018robustness} and \textit{Complexity} \cite{bhatt_evaluating_2020}.\footnote{In this paper, XAI desired properties are italicized and capitalized, whereas XAI evaluation metrics are only capitalized.} However, it should be noted that there is still no agreement on the appellation of these properties, which can be expanded to encompass other desiderata of XAI, as outlined by \textit{Nauta et al.} \cite{nauta_anecdotal_2023} who proposed 12 conceptual properties. However, some of these properties still lack a well-defined quantitative evaluation framework.

To address the aforementioned challenges, we introduce \textbf{BEExAI}, an open-source Python library to \textbf{B}enchmark and to \textbf{E}valuate \textbf{Ex}plainable \textbf{AI} methods. 

\begin{itemize}

\item BEExAI allows the evaluation and comparison of XAI methods on any tabular dataset in a simplified end-to-end pipeline: preprocessing, training, explaining, evaluating and benchmarking.

\item BEExAI allows the computation of 9 evaluation metrics for 8 explainability methods, including LIME, Shapley Values and Integrated Gradients, and supports a wide range of ML models, such as NNs or XGBoost for multi-label classification, binary classification and regression.

\item BEExAI provides optimized implementations of XAI evaluation metrics to assess the quality of explanations according to 3 core XAI desired properties.

\item BEExAI is carefully documented, with detailed examples illustrating various use cases and scenarios to ensure ease of reproducibility and rapid adoption.

\item BEExAI is customizable and flexible: We have provided resources that allow customization of the pipeline at various levels, from data processing to XAI evaluation to enable fast production of benchmark results. The library is, for instance, flexible to integrate new custom explainability methods and evaluation metrics.

\item We carry out a reproducible and rigorous benchmark derived from 50 widely-used datasets that cover regression and classification tasks. This benchmark is the most extensive comparison in Explainable AI, to the best of our knowledge. In addition, we provide a discussion about the peculiarities of each of the three primary tasks in tabular-data related tasks: binary classification, regression, and multi-label classification. We found that, this kind of discussion is generally overlooked in research papers.

\end{itemize}

\noindent
\textbf{Related works:} 

\noindent
With a similar aim to BEExAI, some frameworks have emerged in recent years to compare and evaluate XAI methods. The most popular and pertinent ones include: Captum \cite{kokhlikyan2020captum}, AIX360 \cite{arya2021ai}, XAI-Bench \cite{liu_synthetic_2021}, Quantus \cite{hedstrom2023quantus} and OpenXAI \cite{agarwal2023openxai}:

\begin{itemize}
    \item The Captum library, which is open-sourced by PyTorch, provides access to 22 attribution methods for Deep Learning models, proposing also layerwise and neuronwise attributions in addition to feature-level attributions. It handles various data  formats (images, tabular data, and text). However, there are only two evaluation metrics currently implemented: Infidelity and Sensitivity. While these metrics are essential, they alone do not constitute a complete benchmark, which is not the primary focus of the library.
    
    \item The AIX360 toolkit is an open-source library covering tasks with different data types: images, tabular, text and time series data. The library provides methods for both global and local explanations as well as for post-hoc, ad-hoc, and data-centered XAI approaches. However, only two explainability metrics, Faithfulness and Monotonicity, are available. Like Captum, the focus was not on benchmarking and evaluating XAI methods but on providing multiple implementations for XAI methods.

    \item The XAI-Bench library takes a different approach by benchmarking XAI methods on synthetic data. Five evaluation metrics (Faithfulness, Monotonicity, ROAR, GT-Shapley and Infidelity) are used to compare six XAI methods. However, the use of synthetic data may not always be the best option as it may not accurately represent real-world scenarios \cite{Faber2021WhenCT}. 
    
    \item The Quantus library provides 37 evaluation metrics for XAI methods for neural networks. Similarly to our proposed library, it focuses mainly on evaluation. The library supports general XAI methods of Captum and some layerwise methods as well as explanations from \textit{Zennit} \cite{anders2021software} and \textit{tf\_explain} \cite{Meudec_tf-explain_2021}. However, some of the used evaluation metrics overlap or are different versions of the same metric. Additionally, the library does not provide public benchmarks on reference datasets for comparing supported XAI methods. 

    \item The OpenXAI complements popular metrics from Quantus with other metrics concerning \textit{Fairness} of explanations and adds synthetic data generation for ground truth comparison. The library also provides leaderboards for 6 chosen tabular datasets for comparison on Neural Network and Logistic Regression.
    
\end{itemize}

In Table.\ref{tab:comparison}, we provide statistical comparisons between BEExAI and the aforementioned libraries. In addition, we give in Table.\ref{tab:comparisonprop} an overview of implemented metrics statistics by XAI property. 

\begin{table*}[htbp]
\caption{Comparison of XAI Evaluation libraries}
\label{tab:comparison}
\resizebox{\columnwidth}{!}
{
\begin{tabular}{l|cccccc}
\hline Library & Metrics  &  Models  &  XAI Methods  &  Datasets  &  Benchmark  &  Tasks  \\
\hline  Captum  & 2 & $\infty$ & \textbf{22} & 0 & 0 & reg \& multi\_clf \& bi\_clf \\
 AIX360  & 2 & ? & 20 & 13 & 0 & reg \& multi\_clf \& bi\_clf \\
 XAI-Bench  & 5 & 3 & 6 & 3 & 0 & reg \& multi\_clf \& bi\_clf \\
 Quantus  & \textbf{37} & 1 & 16 & 0 & 0 & multi\_clf \& bi\_clf \\
 OpenXAI  & 22 & 2 & 7 & 8 & 1 & bi\_clf \\
 BEExAI (Ours)  & 9 & 7 & 8 & \textbf{53} & 1 & reg \& multi\_clf \& bi\_clf \\
\hline
\end{tabular}

}
\end{table*}

\begin{table*}[htbp]
\centering
\caption{Number of implemented metrics per key XAI property}
\label{tab:comparisonprop}
\resizebox{0.7\columnwidth}{!}{

\begin{tabular}{l|ccccc}
\hline  Library  &  Faithfulness  &  Robustness  &  Complexity   &  Others  \\
\hline  Captum  & 1 & 1 & 0 & 0\\
 AIX360  & 2 & 0 & 0 & 0\\
 XAI-Bench  & 5 & 0 & 0 & 0\\
 Quantus  & 12 & 12 & 3 & 10\\
 OpenXAI  & 8 & 3 & 0 & 9\\
 BEExAI (Ours)  & 6 & 1 & 2 &  0\\
\hline
\end{tabular}

}
\end{table*}

BEExAI enhances existing libraries by incorporating specific tasks that require deeper exploration due to their inherent intricacy. For instance, regression or multi-label classification present challenges that are not fully addressed by the aforementioned libraries. We also highlight the significant difference in the interpretation of attribution scores between the classification task and regression task and, thus, provide evaluation metrics adapted to each type of task. Moreover, we provide the context required to answer the open question of the baseline’s choice for metrics that require feature ablation. In addition, we focus only on the main evaluation metrics, avoiding redundancy and optimizing computation time. Lastly, we perform a reproducible and rigorous benchmark derived from 50 widely-used datasets that cover both regression and classification tasks.

\section{BEExAI library}\label{sec:beexailib}

BEExAI offers an end-to-end pipeline for practitioners and researchers to generate benchmarks and compare XAI methods using a set of selected metrics. The proposed pipeline consists of 5 main components:

\begin{itemize}
    \item \textbf{Data Preprocessing}: Processing of tabular datasets.
    \item \textbf{Models Training}: Training of ML models.
    \item \textbf{XAI Methods}: Generating explanations (attribution scores) with various XAI methods.
    \item \textbf{Evaluation}: Computing evaluation metrics for XAI methods.
    \item \textbf{Benchmark}: Producing automatic benchmarks to compare multiple datasets, models, and XAI methods.
\end{itemize}

In the following section, we will describe mainly the evaluation component as it is the main focus of this work. 

Other details about the pipeline used for processing, ML models and XAI methods choice can be found in Appendix.\ref{app:pipeline}. Benchmarking procedure will be discussed in section.\ref{sec:benchmarking}.

\subsection{XAI evaluation}\label{sec:evalmetrics}

The literature has explored various metrics and desired properties. However, our purpose is not to include all available metrics, which could introduce unnecessary noise, but to deliver users with a tool that integrates the most commonly used and practical metrics, helping them decide the most suitable XAI method for their use case. Additionally, given the plurality of denominations and formulations attributed to the same metric in the literature, we opt for the most commonly used formulas with low computational complexity to facilitate the understanding of the quantified property and its results.

The evaluation component of BEExAI v0.0.6 provides 9 metrics to quantify 3 main desired properties of explainability: \textit{Faithfulness} \cite{bhatt_evaluating_2020}, \textit{Robustness} \cite{alvarez2018robustness} and \textit{Complexity} \cite{bhatt_evaluating_2020}:

\begin{itemize}
    \item \textbf{\textit{Faithfulness}} metrics are used to measure how accurately XAI methods align with the actual functioning of models in generating predictions. In simpler terms, they determine if the important features selected by XAI methods are actually important and sufficient for making predictions. Most \textit{Faithfulness} metrics achieve this by masking the features in order of their importance scores and analyzing how the model’s prediction changes.

    We mainly focus on \textit{Faithfulness} in this work, an essential property lacking a clear consensus on the best metrics to use. To do this, we implement metrics from various domains of ML, including Natural Language Processing, where we replace the notion of tokens with tabular data features to adapt these metrics to our evaluation framework. However, we have decided not to use metrics from Computer Vision problems as they tend to be too specific and mainly focused on geometric patch ablation.

    \item \textbf{\textit{Robustness}} metrics quantify the degree of stability of the explainability method. For similar instances, we want computed explanations to be also similar. Most \textit{Robustness} metrics involve applying a slight perturbation to a certain instance and computing the relative change in the explainability. The desired outcome is a low change, indicating higher \textit{Robustness}.

    We have chosen to consider fewer metrics for \textit{Robustness} compared to \textit{Faithfulness}, since generating attributions for many perturbed samples tends to be computationally-expensive, especially with perturbation-based methods. In addition, \textit{Robustness} is an easier property to quantify compared to \textit{Faithfulness}, and hence, does not require various evaluation metrics.

    \item \textbf{\textit{Complexity}} metrics quantify the clarity and human interpretability of explanations. We prefer explanations that are sparse rather than dense. This means that we prefer  XAI methods that assign high explainability scores to a small set of features and low scores to the remaining ones. When all features have similar scores, it becomes difficult for users to interpret the results.

    Like \textit{Robustness}, \textit{Complexity} only needs a few metrics to be quantified, as the sparsity of attributions can be often computed using an entropy formula.
    
\end{itemize}

The metrics that are available in BEExAI v0.0.6 are listed in Table.\ref{table:metric}. More details about the metrics can be found in  Appendix.\ref{app:XAIprop}.

\begin{table}
    \centering
    \caption{BEExAI evaluation metrics by XAI property}
    \begin{tabular}{l|l|l}
    \toprule
     \multicolumn{1}{c|}{\textit{\ Complexity \ }}    &  \multicolumn{1}{c|}{\textit{\ Faithfulness \ }} & \multicolumn{1}{c}{\textit{\ Robustness \ }}\\
      \midrule
     \text{\ Complexity \ }    & \text {\ AUC-TP \ } & \text {\ Sensitivity \ }\\
    \text{\ Sparseness \ }    & \text {\ Faithfulness Correlation \ } & \\
        & \text {\ Infidelity \ } & \\
         & \text {\ Monotonicity \ } & \\
         & \text {\ Comprehensiveness \ } & \\
         & \text {\ Sufficiency \ } & \\
    \bottomrule
    \end{tabular}
    \label{table:metric}
\end{table}

\subsection{Choice of baseline}\label{sec:baseline}

Some evaluation metrics require a features’ ablation process, which involves replacing (masking) specific features with a baseline value. Choosing the right baseline is crucial as it needs to represent a neutral reference point for attributions \cite{hameed_based-xai_2022}. One group of baselines consists in sampling feature values from the other instances of the dataset as baselines. Possible methods are:
\begin{itemize}
    \item Uniform or normal distribution,
    \item Opposite: take a sample with a different ground-truth label,
    \item Max distance: select an instance with the maximum L1 distance from the instance of interest while remaining within the features’ distribution.
\end{itemize}

Another alternative is using a \textit{constant} baseline, which is the most common choice. It consists in the replacement of a feature by a constant value, often zero, the mean or the median of the concerned feature.

The benchmarks that follow are based on the zero baseline. This choice is considered as the most appropriate for NNs \cite{ancona_towards_2018} because it leads empirically to a prediction near zero resulting from the multiplication of weights.

Further research is needed to accurately measure the baselines’ choice impact on evaluation metrics’ results and how they align with the intended interpretation of these metrics. This is particularly important when dealing with out-of-distribution instances, which can hinder the production of explanations that are representative of the dataset and the task under consideration \cite{hase2021out}.

It is possible to compute all the mentioned explainability metrics with just one line of code. This requires a curated dataset, a trained model and an explainer as inputs. More parameters can be selected to customize the computation of metrics and get more details about the desired output format. \\

\resizebox{\columnwidth}{!}{
\centering
\begin{lstlisting}[label={lst:listing-python}, language=PythonPlus]
from beexai.evaluate.metrics.get_results import get_all_metrics

get_all_metrics(X_test,model,explainer)
\end{lstlisting}}

\section{Experiments}\label{sec:experiments}
\subsection{Implementation details}\label{sec:implemdetails}
\subsubsection{Main External Libraries Utilized for BEExAI Implementation}\label{sec:externallib}

For the implementation of ML models, we use the scikit-learn library \cite{pedregosa2018scikitlearn}, encompassing Gradient Boosting, Histogram-based Gradient Boosting, XGBoost, Random Forest, Linear Regression and Logistic Regression. Default hyperparameters are retained for these models to ensure easier reproducibility with satisfactory results. The proposed implementation of NNs utilizes PyTorch \cite{paszke2019pytorch}. Hyperparameters’ selection is discussed in Section.\ref{sec:mlmodel}.

Through the BEExAI library, different models can be used to generate benchmarks. This paper provides evaluation results only on XGBoost and NNs models. Considering computational constraints, this choice is made to explore a broader range of XAI methods.

For XAI methods, we use the open-source library Captum \cite{kokhlikyan2020captum} for DeepLift, Integrated Gradients, Saliency, LIME, Shapley Value Sampling and Kernel Shap. Note that other methods like Feature Ablation and Input X Gradient are available in BEExAI but are not included in the following benchmarking results to limit the scope of our analysis.

\subsubsection{Computational details}\label{sec:compute}
BEExAI offers the flexibility of using either GPU or CPU for the entire pipeline, from training the model to the computation of evaluation metrics. Rather than computing the XAI methods for each sample, we suggest an optimized version of the metrics mentioned earlier, which accelerates computation through tensorial operations at various levels. During the production of benchmarks, model weights and attributions can be saved in PyTorch \textit{tensor} format for later reuse, especially for XAI methods with long computation time.

All experiments for the following benchmarks are performed on a device with an AMD EPYC 7742 64-core @ 3.4GHz CPU and multiple 10GB partitions of a single NVIDIA A100 PCIe GPU for multiple random seeds at the same time.

\subsubsection{Sanity check}\label{sec:sanitycheck}
To ensure accurate interpretation of the evaluation metrics’ results and better comparison of different XAI methods, sanity checks are carried out by generating random explainability attributions for each sample within the boundaries of the attributions computed by a given method. We compare, for example, these random attributions with regular attributions in Figure.\ref{absvsrawfig}.

\subsection{Datasets}\label{sec:datasets}

In this section, we present an overview of the datasets employed in our benchmark evaluation. To ensure robustness in handling real-world project tasks, we utilize datasets that are derived from two benchmarks, specifically designed for tabular data.

\begin{itemize}
    \item \textbf{inria-soda/tabular-benchmark} \cite{grinsztajn2022tree} is a selection of datasets from \break OpenML for regression and classification tasks with a specific curation process. This benchmark includes datasets with only numerical features and data with both categorical and numerical features. These datasets are diverse in size and dimensionality, allowing for the generation of explanations and evaluation metrics at different scales.

    We utilized all the datasets present in the benchmark for binary classification. However, for regression, we have excluded ten datasets having an R2 score below 0.3. We choose to restrict our analysis to models with similar performances to limit potential biases. We keep the study of the influence of models’ predictive capability on XAI metric to future work. 

    \item \textbf{OpenML-CC18 Curated Classification} \cite{bischl2017openml} The CC18 benchmark is a suite of 72 classification tasks extracted from other datasets from OpenML with a different curation process than the previous benchmark.
    
    Within these 72 datasets, we selected a subset of 8 datasets with strictly more than two classes to predict, more than 500 training samples and less than 300 features after one-hot encoding. This selection maintains a satisfying variety in size and feature dimensionality, which can lead to more accurate results in multi-classification problems and thus completing the precedent benchmark as we can compare in Table \ref{table:dataset}.

\end{itemize}

\addtolength{\tabcolsep}{1pt}    
\begin{table}[htbp!]
\centering
\caption{Comparison of the two selected benchmarks on their key properties. They cover a wide range of sample size, feature dimensionality and tasks.}
\label{table:dataset} 
\begin{tabular}{lll}
  \toprule
    Benchmark &  {Property} & {Values}\\ 
  \midrule
    \multirow{2}{*}{inria-soda} & {Number of datasets} &45\\
    & {Tasks} & {reg\_num, reg\_cat, clf\_num, clf\_cat}\\
    & {Sample size} & $\in [|2043,752128|]$\\
    & {Features number} & $\in [|3,359|]$\\
    \hline
    \multirow{2}{*}{OpenML-CC18} & {Number of datasets} & 8\\
    & {Tasks} &  multi\_clf\\
    & {Sample size} & $\in [|1178,8793|]$\\
    & {Features number} & $\in [|6,216|]$\\
    & {Number of labels} & 10 \tablefootnote {Except \textit{cmc} dataset with 3 labels}\\
    \bottomrule
\end{tabular} 

\end{table}
\addtolength{\tabcolsep}{-1pt}    

To enhance the accessibility of our benchmark and enable easier usage, our benchmark library includes configurations, models, and precomputed attributions for two specific tasks: a regression task and a multi-label classification task. For regression, we use the Boston Housing dataset, which is derived from the data of Hedonic Prices \cite{harrison1978hedonic}.\footnote{\url{https://www.kaggle.com/code/prasadperera/the-boston-housing-dataset}} For classification, we use the Kickstarter Projects dataset.\footnote{\url{https://www.kaggle.com/datasets/kemical/kickstarter-projects}} These two datasets are an effective starting point to represent each task studied in this work and to facilitate the adaptation of the pipeline to other specific tasks.

\subsection{Benchmarking procedure}\label{sec:benchmarking}

\subsubsection{Data:}\label{sec:data}

The benchmarks were produced on \textbf{OpenML-CC18 Curated Classification} and \textbf{inria-soda/tabular-benchmark} to cover a wide range of tasks with tabular data. We used the provided curated datasets from the two benchmarks without making any additional processing. To reduce the influence of outliers on metrics like Infidelity, which can introduce noise in the calculations, we used QuantileTransformer scaling on the input features. Additionally, we chose to use MinMaxScaler for regression datasets’ target values, restricting outputs to a 0 to 1 range. This prevents unbounded values from affecting the evaluation metrics used to assess explainability.

We split each dataset into 80\% for training and 20\% for testing. As the objective was to produce results on a wide range of tasks without focusing on tuning the hyperparameters of ML predictive models, we did not use a validation set, and directly evaluated our models after training. However, our pipeline allows for creating a validation set and monitoring its performance during training for hyperparameter tuning.

\subsubsection{ML Predictive Models}\label{sec:mlmodel}

We used a three-layer dense NNs with 128 neurons per layer. Between the hidden layers, we added Batch Normalization followed by a 10\% dropout layer. All activations used are Rectified Linear Unit (ReLU), with a softmax layer added at the last layer of the network for classification tasks and a linear layer for regression tasks. The model was trained for 1000 epochs with a learning rate of 1e-3 and a default Adam optimizer. 

For the XGBoost model, we used default hyperparameters defined by scikit-learn, selecting XGBRegressor or XGBClassifier according to the task of interest. 

For classification tasks, the optimization loss is the cross-entropy, whereas for regression, we employ Mean Squared Error (MSE).

\subsubsection{Predictive Models’ evaluation}\label{sec:modeleval}

To evaluate the performance of ML models, we use Accuracy and F1 score for classification tasks, whereas for regression, we measure performance via MSE, Root MSE (RMSE), and R2 score.

\subsubsection{XAI methods’ evaluation}\label{sec:xaieval}

For computational time constraints, we apply stratified sampling to extract 1000 data points from each test-set to ensure that the selected inputs accurately represent the data distribution. We average the results of our XAI metrics benchmarks on five different random seeds for dataset sampling, model training, explainability scores attributions and explainability evaluation. All results are reported with the mean and the corresponding standard deviation (std).

We provide all of our models, attributions and benchmark results for reproducibility and help with further research in the domain.

\section{Task specificity}\label{sec:taskspec}

During our experiments, we realized that to quantify evaluation measures effectively, we need to apply distinct treatments to the generated explainability attributions and evaluations metrics based on the type of task of interest (binary classification, multi-label classification, or regression). To the best of our knowledge, there is a lack in previous research addressing this matter.

\subsection{Classification}
To compute the attributions, we select, for each instance, the label with the highest predicted probability as the target. We believe that this approach provides a more accurate portrayal of the model’s predictive capability. Afterwards, we calculate the evaluation metrics using the same chosen targets.
    
An alternate solution that we discarded is to choose the ground-truth label. Indeed, in real-world scenarios, we do not have access to ground-truth labels, making this method impractical.

We have also experimented with other alternatives. One approach we tried was to compute attributions and evaluation measures for each label, then take the mean of these values across all labels to obtain a global score for each instance. However, this approach was computationally expensive, especially for multi-classification tasks, and it did not yield accurate results as the predicted probabilities for classes that were not predicted might not accurately reflect the model’s predictive capability. We also considered the approach of summing on the absolute values of the attributions for each label before calculating the explainability metrics. The assumption behind this approach was that we were only concerned with the relative importance of each feature. However, this possibility was discarded as it was not accounting for the fact that the attributions’ sign has a significant role in classification tasks.

We made our choice of treatment according to our results, but we emphasize the need for in-depth research to analyze the best method that ensures consistent comparisons between XAI methods. 
    
\subsection{Regression} 
\subsubsection{Attributions} We consider the absolute values of the generated attributions instead of raw values. This choice is motivated by the specificity of regression tasks compared to classification ones. Indeed, in regression tasks, the goal is to predict a scalar value, not a probability as in classification tasks. Hence, the target value can be approached by positive and negative contributions from the input features. Negative contributions can offset positive ones, leading to faster convergence toward the target. Therefore, we prioritize the features with the highest contributions, regardless of whether they are positive or negative. We found that retaining the signs of attributions led to less satisfactory results for metrics like Comprehensiveness and Sufficiency.

\begin{figure}[htbp]
\centering
\begin{minipage}[b]{0.48\textwidth}
\centering
\includegraphics[width=\textwidth]{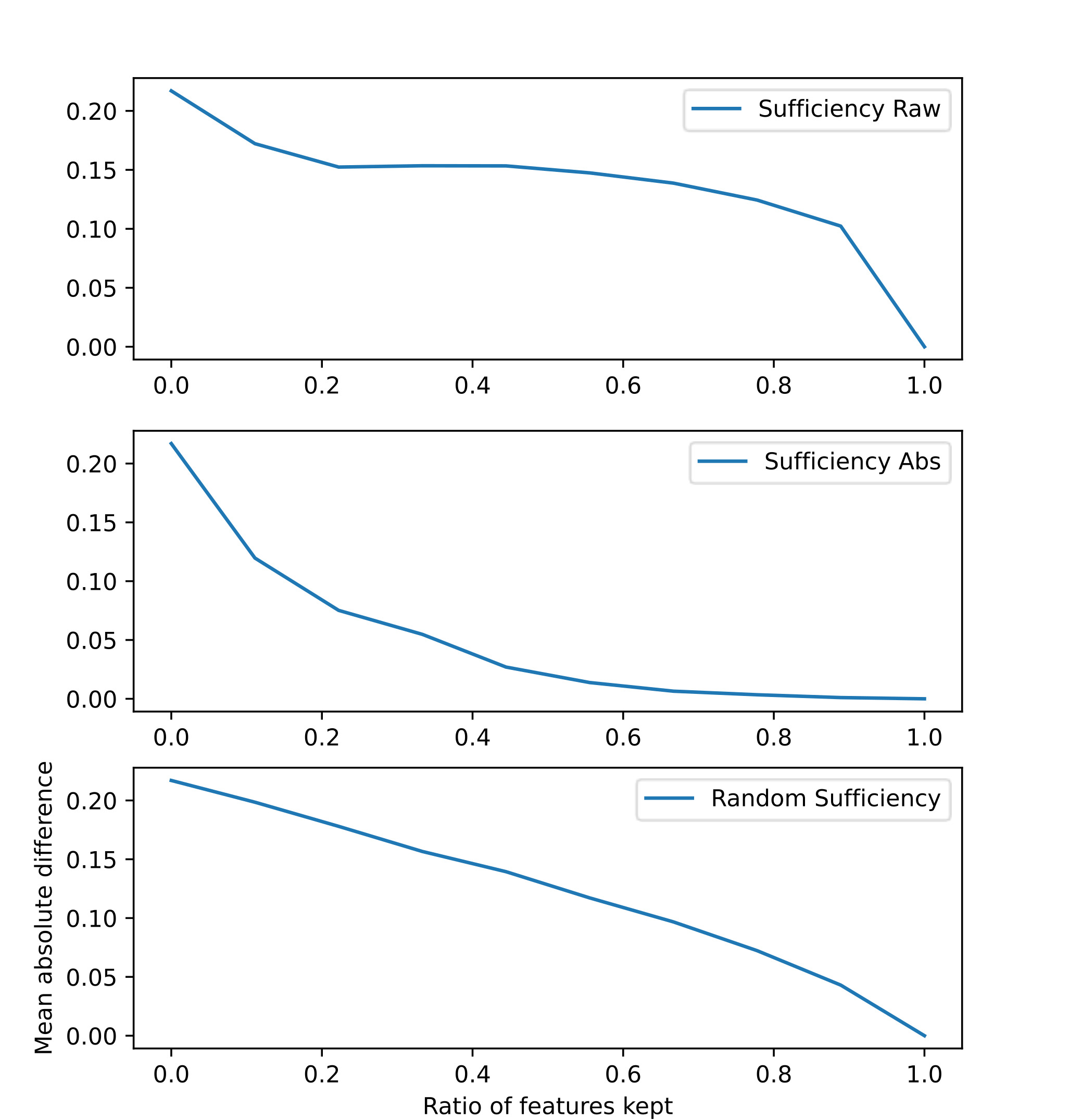}
\end{minipage}
\centering
\hfill
\begin{minipage}[b]{0.48\textwidth}
\centering
\includegraphics[width=\textwidth]{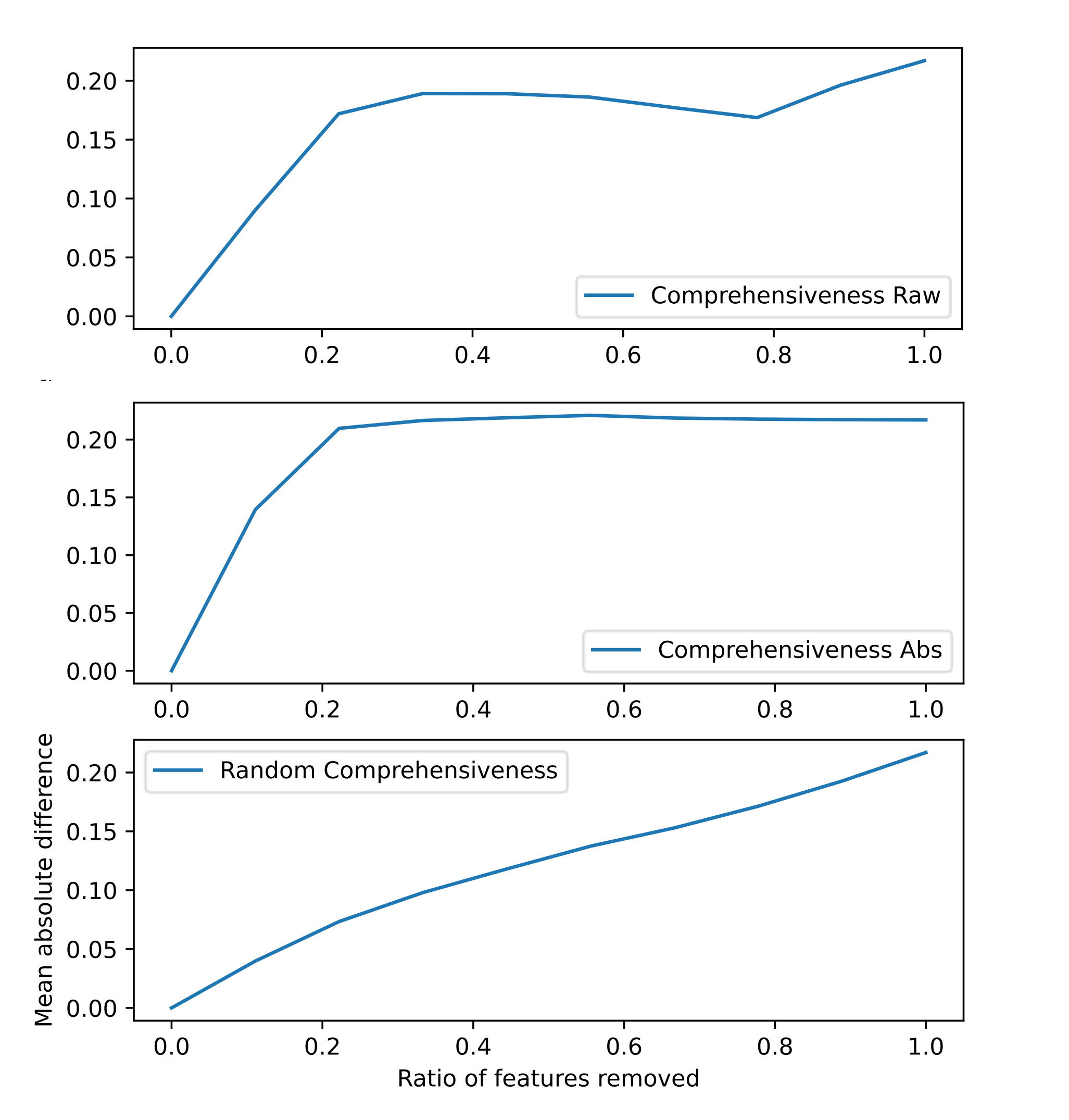}
\end{minipage}
\caption{Sufficiency (left) and Comprehensiveness (right) curves on \textit{Diamonds} dataset for regression task. These curves represent the average values obtained from 1000 samples with XGBoost model and ShapleyValueSampling method.
Sufficiency (Comprehensiveness) values refer to the absolute difference between the ML model’s predicted value when all features are used (removed) and the predicted values of the same model when the features are successively removed (added) in an ascending order based on their corresponding explainability attributions.\\
Top: Sufficiency and Comprehensiveness values based on absolute values of attributions. Middle: Sufficiency and Comprehensiveness values based on raw attributions. Bottom: Sufficiency and Comprehensiveness for random attributions.
}
\label{absvsrawfig}
\end{figure}

Figure.\ref{absvsrawfig} shows an example of how using absolute values of attributions instead of raw attributions affects Sufficiency and Comprehensiveness metrics. For Sufficiency, an expected outcome is observed with the application of absolute values. The most important features that are ablated first have the greatest impact on the metric value. As we remove the less important features, the impact decreases. This means that the most important features are sufficient to explain the model’s predictions, while the less important features are less essential. However, when using raw attributions, the values of Sufficiency metric become less obvious. The resulting trend has three phases. First, removing features with the highest positive attributions noticeably decreases the metric value. Second, as less important features are removed, the metric changes slightly. Finally, removing features with the highest negative attributions causes another decrease in the metric’s value.

\subsubsection{Metrics} We choose to adjust specific metrics (Faithfulness Correlation, Comprehensiveness, Sufficiency, and Monotonicity) that consider the difference between the model’s prediction on a sample and the same model’s prediction of a perturbed version by taking the absolute value of this difference instead of the signed difference. Indeed, the interpretation of the difference’s sign is different if we are on a regression or a classification scenario. In the case of classification, the prediction is represented as a probability for the label of interest. Hence, when masking the first most important features, we expect a positive difference that represents a probability’s degradation of the label of interest (the predicted label). However, in the case of regression, when masking the first most important features, this degradation can be either positive or negative because the prediction represents a scalar value, and thus, the sign of the difference has not the same significance as it does in classification tasks.

\section{Results}\label{sec:results}

In this section, we present the XAI methods’ evaluation results of two ML models: NNs and XGBoost. Indeed, we found that the two methods perform similarly in terms of accuracy for classification and R2 score for regression tasks.\footnote{It is worth noting that we did not explore hyperparameters for XGBoost models as they were already performing well, and specific model tuning was not the main focus of this work.} This similarity of performance is an excellent point to focus on comparison between XAI methods.

\begin{table}[!htbp]
\caption{Evaluation metrics’ results on the \textit{Diamonds} dataset within a regression task scenario, calculated across five random seeds. \small{\textit{Sensitivity} metric values have been scaled up by a factor 100 due to small original values. The last row provides NNs and XGBoost predictive performance (R2).}}
\centering
\resizebox{\columnwidth}{!}{
\sisetup{detect-all}
\NewDocumentCommand{\B}{}{\fontseries{b}\selectfont}
\small
\begin{tabular}{
  @{}
  l
  *{1}{S[table-format=-1]}|
  *{1}{S[table-format=-1]}
  *{1}{S[table-format=-1]}
  *{1}{S[table-format=-1]}
  *{1}{S[table-format=-1]}
  *{1}{S[table-format=-1]}
  *{1}{S[table-format=-1]}
  @{}
}
\toprule
Methods & & {LIME} & {Shap} & {KShap} & {DL} & {IG} & {Saliency}\\
\midrule
\multirow{2}{*}{Faith \scalebox{.8}{$\nearrow$}}& NN & {$0.728_{\mypm .022}$} & {$\mathbf{0.884}_{\mypm .017}$} & {$0.773_{\mypm .033}$} & {$0.784_{\mypm .042}$} & {$0.838_{\mypm .029}$} & {$0.302_{\mypm .034}$}\\
& XGB\  & {$0.837_{\mypm .027}$} & {$\mathbf{0.911}_{\mypm .031}$} & {$0.87_{\mypm .024}$}\\
\hline
\multirow{2}{*}{Inf \scalebox{.8}{$\searrow$}}& NN & {$0.101_{\mypm .004}$} & {$0.102_{\mypm .004}$} & {$0.102_{\mypm .004}$} & {$0.107_{\mypm .004}$} & {$0.1_{\mypm .003}$} & {$\mathbf{0.098}_{\mypm .003}$}\\
& XGB\  & {$\mathbf{\ 0.108}_{\mypm .005}$} & {$0.112_{\mypm .005}$} & {$0.113_{\mypm .005}$}\\
\hline
\multirow{2}{*}{Sens \scalebox{.8}{$\searrow$}}& NN & {$0.076_{\mypm .001}$} & {$0.068_{\mypm .001}$} & {$0.069_{\mypm .001}$} & {$0.073_{\mypm .002}$} & {$0.068_{\mypm .001}$} & {$\mathbf{0.033}_{\mypm .002}$}\\
& XGB\  & {$0.068_{\mypm .003}$} & {$\mathbf{0.067}_{\mypm .002}$} & {$0.068_{\mypm .002}$}\\
\hline
\multirow{2}{*}{Compr \scalebox{.8}{$\nearrow$}}& NN & {$0.101_{\mypm .005}$} & {$\mathbf{0.11}_{\mypm .004}$} & {$0.103_{\mypm .004}$} & {$0.086_{\mypm .004}$} & {$0.105_{\mypm .005}$} & {$0.077_{\mypm .006}$}\\
& XGB\  & {$0.197_{\mypm .009}$} & {$\mathbf{0.199}_{\mypm .01}$} & {$0.195_{\mypm .01}$}\\
\hline
\multirow{2}{*}{Suff \scalebox{.8}{$\searrow$}}& NN & {$0.078_{\mypm .003}$} & {$\mathbf{0.068}_{\mypm .003}$} & {$0.074_{\mypm .003}$} & {$0.098_{\mypm .005}$} & {$\mathbf{0.068}_{\mypm .003}$} & {$0.11_{\mypm .008}$}\\
& XGB\  & {$0.079_{\mypm .009}$} & {$\mathbf{0.075}_{\mypm .012}$} & {$0.08_{\mypm .011}$}\\
\hline
\multirow{2}{*}{Mono \scalebox{.8}{$\nearrow$}}& NN & {$0.7_{\mypm .061}$} & {$0.825_{\mypm .061}$} & {$0.775_{\mypm .05}$} & {$\mathbf{0.975}_{\mypm .05}$} & {$0.825_{\mypm .061}$} & {$0.875_{\mypm .0}$}\\
& XGB\  & {$0.675_{\mypm .061}$} & {$\mathbf{0.9}_{\mypm .05}$} & {$0.825_{\mypm .061}$}\\
\hline
\multirow{2}{*}{AUCTP \scalebox{.8}{$\searrow$}}& NN & {$0.232_{\mypm .012}$} & {$0.204_{\mypm .008}$} & {$0.223_{\mypm .01}$} & {$0.243_{\mypm .008}$} & {$\mathbf{0.2}_{\mypm .01}$} & {$0.238_{\mypm .009}$}\\
& XGB\  & {$0.322_{\mypm .069}$} & {$\mathbf{0.321}_{\mypm .101}$} & {$0.348_{\mypm .088}$}\\
\hline
\multirow{2}{*}{Compl \scalebox{.8}{$\searrow$}}& NN & {$\mathbf{\ 0.081}_{\mypm .004}$} & {$0.194_{\mypm .001}$} & {$0.205_{\mypm .001}$} & {$0.194_{\mypm .001}$} & {$0.197_{\mypm .001}$} & {$0.204_{\mypm .001}$}\\
& XGB\  & {$\mathbf{\ 0.083}_{\mypm .003}$} & {$0.164_{\mypm .003}$} & {$0.182_{\mypm .003}$}\\
\hline
\multirow{2}{*}{Spar \scalebox{.8}{$\nearrow$}}& NN & {$\mathbf{0.79}_{\mypm .008}$} & {$0.479_{\mypm .006}$} & {$0.436_{\mypm .003}$} & {$0.482_{\mypm .008}$} & {$0.465_{\mypm .006}$} & {$0.442_{\mypm .007}$}\\
& XGB\  & {$\mathbf{\ 0.787}_{\mypm .007}$} & {$0.613_{\mypm .012}$} & {$0.544_{\mypm .014}$}\\
\hline
\multirow{2}{*}{R2 \scalebox{.8}{$\nearrow$}}& NN & {$0.979_{\mypm .002}$}\\
& XGB\  & {$0.993_{\mypm .001}$}\\
\bottomrule
\end{tabular}
}
\label{tab:reg_cat_diamonds}
\end{table}

To facilitate the interpretation of our results, we distinguish between two categories. Metrics where high values are desired are marked with $\nearrow$, signifying a goal value of 1 since all metrics are normalized. Conversely, metrics where low values are desired are denoted with $\searrow$, indicating a goal value of 0.\footnote{Note that due to the small values of the Sensitivity metric, the reported results of this metric are scaled by a factor of 100.}

\begin{figure}[!htbp]
\centering
\includegraphics[width=.8\textwidth,keepaspectratio]{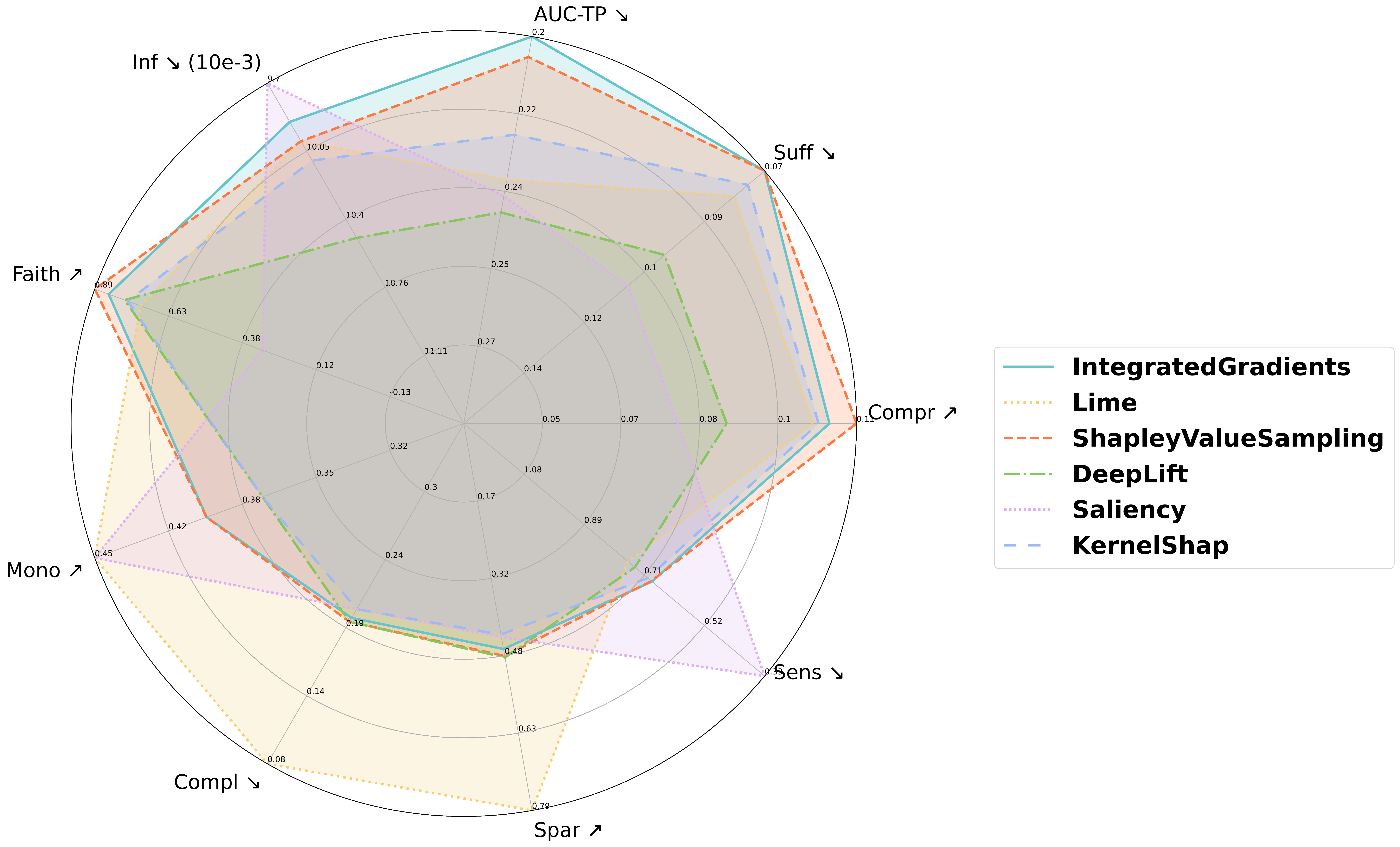}
\caption{Radar plot for explainability metrics values for the \textit{Diamonds} dataset within a regression task scenario using a NNs model. \small{\textit{Sensitivity} metric values have been scaled up by a factor of 100.}}
\label{tab:radar}
\end{figure}

Table.\ref{tab:reg_cat_diamonds} provides the results of the XAI evaluation metrics for the \textit{Diamonds} dataset within a regression task scenario. This dataset includes 43,152 samples and 9 features, with numerical and categorical variables. More results regarding the other tasks and datasets are available at \url{https://github.com/SquareResearchCenter-AI/BEExAI/tree/main/benchmark_results}. Furthermore, for a simpler representation and comparison of different XAI methods, we visualize the results for this dataset through a radar chart in Figure.\ref{tab:radar}.

\begin{figure}[htbp]
\centering
\includegraphics[width=\textwidth,keepaspectratio]{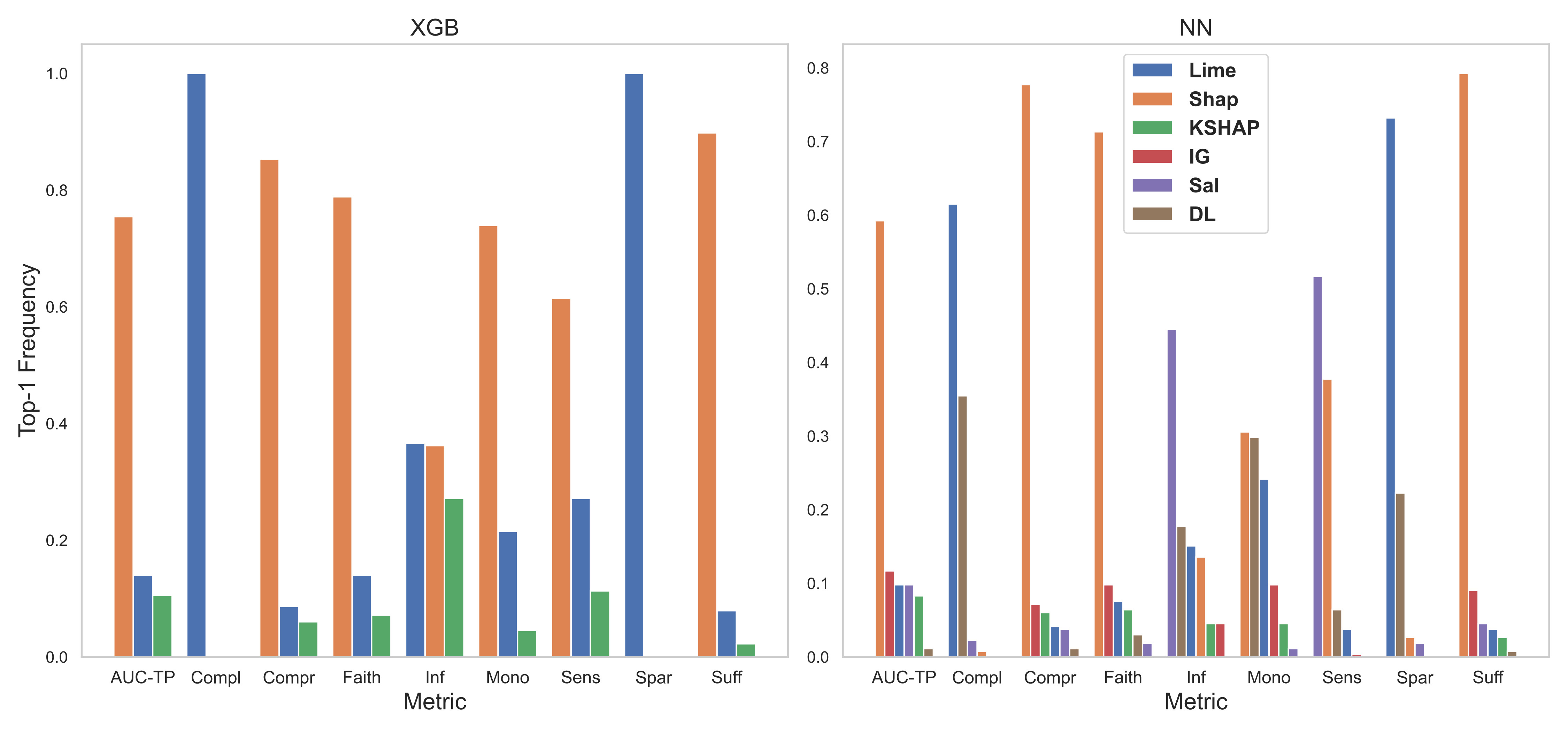}
\caption{Comparative performance of XAI methods: Frequency (in terms of percentage) of top-1 rankings across evaluation metrics for XGBoost (left) and NNs (right). \small{XAI methods with missing bars indicate a top-1 ranking frequency close to 0\%}}

\label{fig:topone}
\end{figure}

\begin{figure}[htbp]
\centering
\includegraphics[width=\textwidth,keepaspectratio]{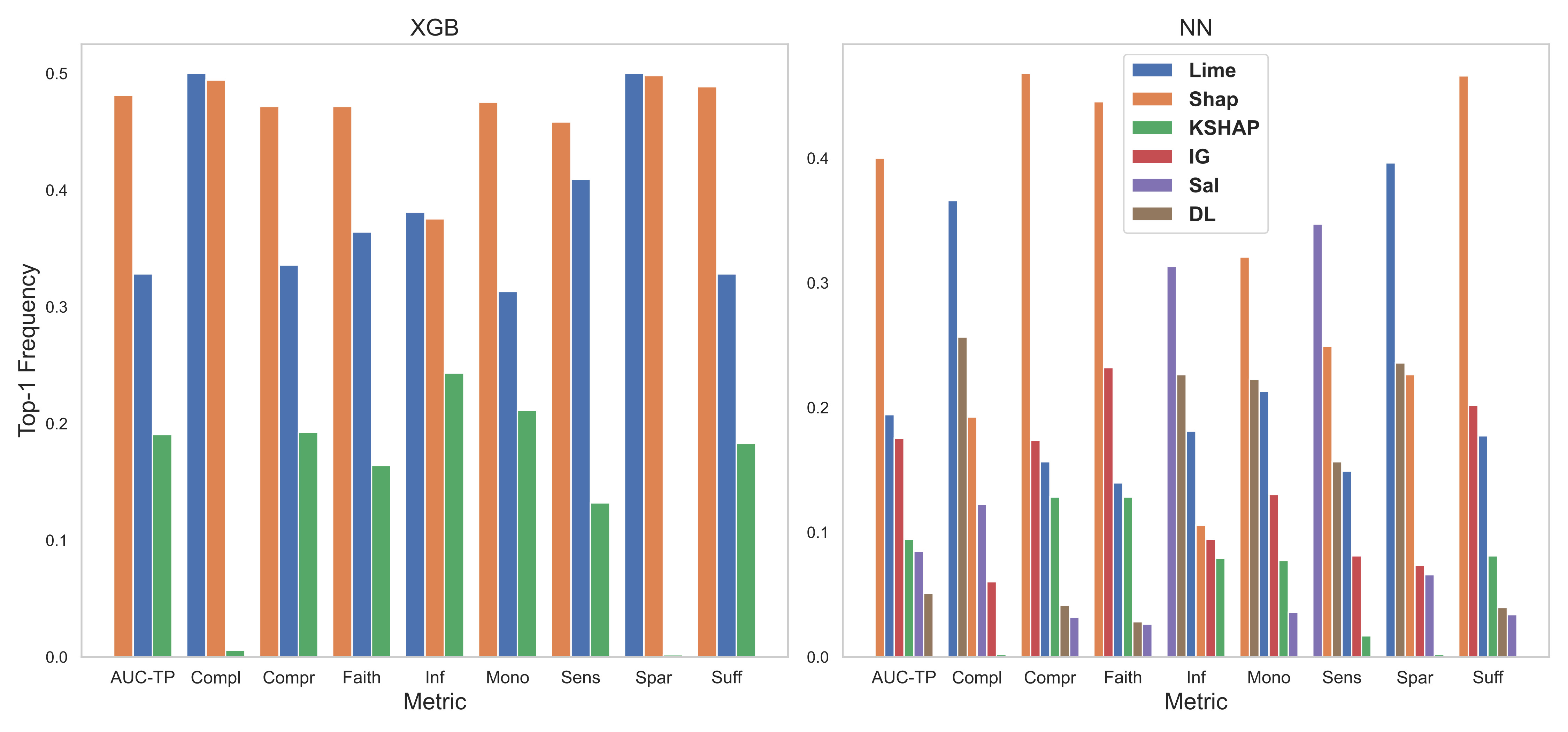}
\caption{Comparative performance of XAI methods: Frequency (in terms of percentage) of a combined count of top-1 and top-2
rankings across evaluation metrics for XGBoost (left) and NNs (right).}
\label{fig:toponetwo}
\end{figure}

In addition, to have a comprehensive understanding and a global comparison between the different XAI methods across various datasets, we propose comparing them based on how frequently each method is ranked first among the others for each metric for both XGBoost and NNs.
Figure \ref{fig:topone} illustrates the results of this measure in terms of percentage. In addition, as metric values might be close for different XAI methods, we include the combined count of top-1 and top-2 rankings for additional comparison in Figure \ref{fig:toponetwo}.

In the following, we discuss the obtained results in terms of the studied XAI desired properties: \textit{Faithfulness}, \textit{Robustness} and \textit{Complexity}.

\textbf{\textit{Faithfulness}} is quantified by computing Faithfulness Correlation, Infidelity, Comprehensiveness, Sufficiency, Monotonicity and AUC-TP. On \textit{Diamonds} dataset, from the results provided in Table.\ref{tab:reg_cat_diamonds}, we found that Shapley Value Sampling (SHAP) generates more faithful explanations, especially on Comprehensiveness, Sufficiency, AUCTP, and Faithfulness Correlation. However, some other methods like Integrated Gradients perform slightly better on certain tasks, and the differences in the values obtained can be negligible in some cases.

\noindent
From a comprehensive perspective across various datasets and tasks, the \textit{Diamonds} dataset’s results are affirmed by counting the number of times SHAP is ranked first in terms of the four metrics mentioned above compared to the other XAI methods as shown in Figure.\ref{fig:topone}. For NNs models, SHAP achieves over 59\% first ranking on the four metrics, while other XAI methods are ranked first at most 12\% of datasets in both regression and classification. 
We observe similar results for XGBoost models when comparing SHAP to LIME and KernelShap XAI methods. 

\noindent
Integrated Gradients and LIME also yield good results for the same metrics when considering the second ranking in Figure.\ref{fig:toponetwo}. 

\noindent
Concerning the Monotonicity metric, we found that SHAP yields the highest results for XGB models, achieving over 70\% first ranking. The top ranking for NNs is almost equally divided between SHAP, DeepLift, and LIME, with 33\%, 32\%, and 28\%, respectively

\noindent
Concerning the Infidelity metric, we found that Saliency explainability method achieves better results than other methods on NNs models (first ranked in 45\% of cases). For XGBoost-based models, LIME and SHAP have the same first-place ranking rates (38\%).

\textbf{\textit{Robustness}} is evaluated using the Sensitivity metric. On \textit{Diamonds} dataset (Table.\ref{tab:reg_cat_diamonds}), Saliency outperforms significantly the other XAI methods with a value more than 200\% lower on NNs models. For XGBoost, the three experimented XAI methods yield comparable results.  

\noindent
Globally in Figures \ref{fig:topone} and \ref{fig:toponetwo}, we observe that SHAP and Saliency provide the best results with, respectively, 38\% and 52\% of top-1 ranking with NNs, while SHAP attains 62\% top-1 ranking with XGBoost. Similar results are observed when considering the combined frequency of top-1/top-2 rankings.

\textbf{\textit{Complexity}:} LIME outperforms other XAI methods with scores that are up to twice as good in some cases. This observation was expected since a sparse linear model is used as a surrogate of the \textit{black-box} model in the LIME framework. Regarding ranking, it achieves 73\% and 62\% top-1 rankings for, respectively, Sparseness and Complexity metrics for NNs, and 100\% for both metrics with XGBoost. 
Despite this, DeepLift still gives good results for NNs with 22\% and 35\% of top-1 ranking (resp. Sparseness and Complexity), while other XAI methods are ranked first less than 3\% of the time across all datasets. SHAP is often ranked 2nd in terms of \textit{Complexity}, closely behind LIME and DeepLift.
\smallbreak
From our analysis, we can conclude that BEExAI can help users choose which XAI methods might be the most suitable given the model and the task at-hand. For instance, our results have shown that ShapleyValueSampling generally outperforms others in terms of Faithfulness properties. Saliency has been shown to be a robust option compared to the other methods in our benchmark. LIME and DeepLift can also be two pertinent choices if practitioners desire less complex XAI methods. We strongly advise adapting the choice of an XAI method depending on the task tackled, the model chosen and the desired property. We hope that BEExAI and our benchmark's results will facilitate this process.

\section{Conclusion}\label{sec:conclusion}

We have presented BEExAI, an open-source library that can be used for benchmarking and evaluating post-hoc explainability attribution methods. This library aims to establish a standard evaluation of XAI methods through quantitative analysis. It provides a simplified and reproducible end-to-end pipeline that allows users to evaluate, compare, and benchmark XAI methods on any tabular dataset, covering a wide range of ML models, XAI methods, and metrics. Furthermore, we provide the largest benchmark available to date for three fundamental tasks in tabular-data problems: regression, binary classification, and multi-label classification. We hope that BEExAI will contribute to the development of XAI in the future and serve as a foundation for the research community.

Throughout this paper, we have shared insightful reflections on the challenges that arise when studying XAI evaluation. First, we have discussed the challenges raised for each task type and their impact at different stages of the evaluation process. Second, we have delved into the baseline choice for evaluation metrics that involve a feature ablation process. Third, we have proposed a sanity check that compares random explanations with explanations of XAI methods to assess the quality of generated explanations. However, although we have discussed and proposed solutions to these challenges, we believe that in-depth research is needed to gain a thorough understanding. 

To complete BEExAI, we propose a set of perspectives that could significantly enhance its capabilities.
Regarding the data, expanding our library to include NLP and Computer Vision pipelines would be a welcome extension. In addition, it would be interesting to consider a more extensive evaluation of XAI methods (1000 data-points are used for this paper’s results).
Regarding the models, we hope that future work addresses the influence of models’ predictive capability on explainers’ performance, especially when two ML models have significant discrepancies in performance.
For explainers, future work could focus on integrating other types of explanations, such as rule-based methods, into BEExAI.

Finally, we believe that quantitative analysis should be associated with a human assessment of the results, i.e., by studying the \textit{Plausibility} of explanations.

\begin{credits}
\subsubsection{Acknowledgments.}\label{sec:acknowledgements}
This work is supported by Square Management which we thank for the funds allocated in computing resources. We thank Pr. Alexandre Allauzen and the Square Management consultants that contributed to this research in particular Antoine Boulinguez and Sofian Mzali. We also thank our anonymous reviewers for their insightful comments and questions that helped us improve the quality of this paper.
\subsubsection{Disclosure of Interests.}
The authors have no competing interests to declare that are relevant to the content of this article.
\end{credits}
\bibliographystyle{splncs04}  
\bibliography{references}

\section*{Appendix\label{appendix}}

BEExAI includes a complete pipeline to facilitate the production of benchmarks for XAI, for which we detail the main components in this first appendix.

We also provide more insights on the chosen explainability metrics with their properties and specific implementation choices made for our benchmarks.

\renewcommand{\thesubsection}{\Alph{subsection}}
\subsection{BEExAI other components}\label{app:pipeline}
\subsubsection{Data processing}\label{sec:dataprocessing}
BEExAI provides a pre-built preprocessing component for tabular data. Its main purpose is to simplify the production of large-scale benchmarks. Different options are proposed to prepare datasets for models' training:

\begin{itemize}

\item \textbf{Automated DateTime features processing:} Automatically convert features with DateTime format by extracting key time information such as year, month, day, and hour.

\item \textbf{Creation of new features:} Adding new features based on elementary operations on other features. This can help to provide more context and insights for the model and improve its accuracy.

\item \textbf{Removing non-relevant samples:} Delete rows having a feature with a specific value.

\item \textbf{Correlated features:} To avoid redundancy and overfitting, automatically keep only features with a cross-correlation lower than a given threshold.

\item \textbf{Standardization:} Numerical input features and regression target are automatically standardized to warrant an accurate comparison. This includes Standard, Min-Max, Max-Abs, Robust scaler and Quantile transformation with uniform or Gaussian distribution.

\item  \textbf{Categorical features encoding:} Ordinal encoding or One-Hot encoding.

\end{itemize}

These steps allow processing a custom tabular dataset associated with a configuration file into a curated dataset, and splits for training and evaluation.\\

It should be noted that using this component is optional. We invite users to provide their own processed datasets, if available. Indeed, the main objective of this library is not to maximize models' performance. Hence, this component provides a good starting point for simple datasets to focus more on the XAI, but there is room for further improvement with more preprocessing options.  

\subsubsection{Model Training}\label{sec:training}

BEExAI provides an easy and automatic training component for multi-label classification, binary classification and regression. It allows using the most commonly used ML models, such as XGBoost, Random Forest, Decision Tree, Gradient Boosting, and Logistic and Linear Regression.

We also provide an implementation of fully-connected NNs with linear layers and ReLU activations. Users can adjust hyperparameters, such as hidden layers' number and regularization. We leave the exploration of more intricate NNs architectures to future studies, e.g., data-types like Natural Language Processing (NLP), where the use of Large Language Models add intricacy, or in Computer Vision, where Convolutional NNs are commonly used.

A model can easily be instantiated and trained with a  \textit{Model} class and a preprocessed training set. Custom hyperparameters can be provided to enhance the model beyond the default settings to make it more suitable for the specific dataset at hand. We also provide training options with K-fold cross-validation and GridSearch to tune previously mentioned scikit-learn models' hyperparameters.

\subsubsection{Explainers}\label{sec:explainer}

The explainers' component provides access to the most popular XAI methods. Note that, as aforementioned, our focus lies exclusively on local post-hoc feature attribution methods. Rule-based approaches such as Anchors and example-based methods like Counterfactuals are not within the scope of this work. These are reserved for future research and independent benchmarking tools, as we believe that distinct evaluation frameworks are necessary for each type or format of explanation.

The first category of XAI methods available in BEExAI v0.0.6 are \break perturbation-based methods that modify the input data and measure the changes in the model's predictive ability. These XAI methods estimate the importance of features for a single instance by comparing the difference between the model's output when using the original instance and the model's output when a perturbed version of the instance of interest is used. To obtain a good approximation for the computed attributions, these methods often repeat this process with several perturbations.

Most perturbation-based methods are, by definition, model-agnostic, although some combinations with gradient-based methods such as TreeSHAP \cite{lundberg2018consistent} or DeepSHAP \cite{lundberg2017unified} have emerged in recent years. This allows attribution computation for a wide range of ML models. However, this type of method has a higher computational cost due to the iterative process on multiple forwards passes with several perturbed samples. Additionally, these methods can be sensitive to the choice of the perturbation function. 

We consider the following XAI methods from different public repositories:
\begin{itemize}
    \item LIME  (Local Interpretable Model-Agnostic Explanations) \cite{ribeiro2016why}
    \item Shapley Value Sampling \cite{trumbelj2010AnEE}
    \item Kernel SHAP \cite{lundberg2017unified}
    \item Feature Ablation, a particular case of
    \cite{zeiler2014visualizing}
\end{itemize}

The second category of XAI methods considered in BEExAI v0.0.6 are gradient-based methods specific to models with gradient descent optimization, like Deep NNs. These methods use gradient-based computations to calculate the attributions of the output with respect to the input features by backpropagation.

Gradient-based methods require less computational resources since they only need to make one or a few backwards passes. These methods are specifically designed for NNs, making them more conceptually linked to the underlying architecture, potentially increasing trust in their generated explanations. However, they do not apply to ML models commonly used for tabular-data-related tasks like Gradient boosting, limiting their application scope. Additionally, gradient-based methods often rely on the choice of a baseline, which can significantly vary across different tasks.\footnote{Here baseline means a reference for gradient-based methods like Integrated Gradients or DeepLift. It should not be confused with baseline used in explainability metrics that we introduce in \ref{sec:baseline}.} 

There are various gradient-based methods. We consider the most popular ones:
\begin{itemize}
    \item Saliency \cite{simonyan2013deep}
    \item Integrated Gradients \cite{sundararajan2017axiomatic}
    \item DeepLift (Deep Learning Important FeaTures) \cite{shrikumar2019learning}
    \item Input X Gradient \cite{ancona_towards_2018}
\end{itemize}

With different Explainer classes, one can obtain explanations for a trained model in just a few lines of code. Users can add their new custom explainers.

\subsection{XAI metrics}\label{app:XAIprop}

\noindent
\textbf{Complexity} \cite{bhatt_evaluating_2020} computes the Shannon entropy of features' fractional contributions. The fractional contribution of a feature $i$, denoted $\mathbb{P}_{g}(i)$, is defined as: $\mathbb{P}_{g}(i) = \frac{|g(f,x)_{i}|}{\sum_{s \in S} |g(f,x)_{s}|}$,
where $S$ represents the features set, $g$ is the explainability function, $f$ is the predictor and $x$ a specific instance. Complexity, denoted $\mu_{C} (f,g;x)$, is defined as: 

\begin{equation}
\mu_{C} (f,g;x) = \mathbb{E}_{i}[-ln(\mathbb{P}_{g})] = - \sum_{s \in S} \mathbb{P}_{g}(i) ln(\mathbb{P}_{g}(i)).
\end{equation}

The objective is to minimize complexity, with a set of features ideally having high contributions and the others approaching 0. On the contrary, uniformly distributed attributions among features result in high complexity. In our implementation, we standardize complexity to get values ranging from 0 to 1. \\

\noindent
\textbf{Sparseness} \cite{chalasani_concise_2020} is another way of quantifying the concentration of explanations on specific features. Given an attribution vector $v = [v_{1},v_{2},...,v_{d}]$ with non-negative values sorted in non-decreasing order, Sparseness is defined with the Gini Index: 

\begin{equation}
G(v) = 1 - 2\sum_{k=1}^{d} \frac{v_{k}}{||v||_{1}}(\frac{d-k+0.5}{d}). 
\end{equation}

By definition, $\forall v \in (\mathbb{R}^{+})^{D}, G(v) \in [0,1]$. Ideally, one $v_{k} > 0$ with others being 0 leads to $G(v)=1$. Uniformly distributed attributions with nonzero values result $G(v)=0$, which is undesirable. \\

\noindent
\textbf{Faithfulness correlation} \cite{bhatt_evaluating_2020} quantifies the correlation between the sum of attributions of $x_{S}$ (where $x_{S}$ denotes $x$ with a subset $S$ of features replaced by a baseline value) and the difference in model predictions between $x_{S}$ and the original input $x$. Given a predictor function $f$, an explanation function $g$, an input instance $x$, and a feature subset $S$, the faithfulness correlation is defined as: 
\begin{equation}
\mu_{F}(f,g;x) = \underset{S \in \left( \begin{array}{c}
|d| \ |S| \end{array} \right)}{corr} (\sum_{i \in S} g(f,x)_{i},f(x)-f(x_{S})). 
\end{equation}

Note that not all subsets within $\left( \begin{array}{c}|d| \ |S| \end{array} \right)$ are explored and the size of subsets $|S|$ is fixed. By definition, the correlation values range from -1 to 1. Ideally, $\mu_{F}=1$.  In our experiments, we set the subset size $|S|$ as a percentage of the total number of features, with a default value of 20\%, which we found yields sufficiently discriminating results between different explainability methods. We iterate this process 20 times, which we consider a good trade-off between computation time and approximation quality.\\

\noindent
\textbf{Area Under the Threshold Performance Curve} \cite{atanasova_diagnostic_2020} computes the AUC of the curve 
$(q,P(f(x) - f(x_{q\%}))$, where $P$ represents the performance measure of the task of interest and $x_{q\%}$ denotes the input with the $q\%$ most important features (according to the explainability method under consideration) replaced by a baseline, with $q \in [0,10,...,100]$. We expect the performance curve will notably decrease after removing the most important features, hence the objective is to minimize the AUC. We standardize the final results by the number of features in the dataset to ensure comparability across different tasks. \\

\noindent
\textbf{Comprehensiveness} \cite{deyoung2019eraser} quantifies the impact of replacing the $q\%$ most important features by a baseline value. It is defined as:

\begin{equation}
C(x,q) = f(x) - f(x_{q\%})
\end{equation}

\noindent
where $x_{q\%}$ represents the input $x$ with the $q\%$ most important features replaced by a baseline. We expect that as we replace features in a non-ascending order of their attributions, the metric value will increase with the augmentation of $q\%$, the percentage of replaced features. The most important features that are ablated first have the greatest impact on the metric value. As we remove the less important features, the impact decreases.\\

\noindent
\textbf{Sufficiency} \cite{deyoung2019eraser} measures the impact of including the most important features to a baseline on the model's output. It is complementary to comprehensiveness and is defined as:

\begin{equation}
S(x,r) = f(x) - f(x_{r\%})
\end{equation}

\noindent
where $x_{r\%}$ represents the input $x$ with only $r\%$ of the most important features added starting from a baseline. Ideally, as we incorporate the most important features in a non-ascending order of their attributions, we expect the prediction to converge towards the prediction when the original input is used.

\noindent
For both Comprehensiveness and Sufficiency, we use a ratio of $30\%$ for $q$ and $r$. \\

\noindent
\textbf{Monotonicity} \cite{liu_synthetic_2021} compares the
marginal impact of a set of features to their corresponding weights. Given an ensemble $S_{i}$ comprising the $i$ most important attributions, an instance $x$ and a predictive function $f$, Monotonicity is defined as: 

\begin{equation}
m = \frac{1}{D-1} \sum_{i=0}^{D-1} \mathbb{I}_{|\delta_{i}| \le |\delta_{i+1}|}
\end{equation}

\noindent
with $\delta_{i} = f(x_{S_{i+1}}) - f(x_{S_{i}})$. An ideal Monotonicity value is 1, indicating that each subsequent feature provides better marginal improvement than a less important feature.\\

\noindent
\textbf{Infidelity} \cite{yeh2019fidelity} quantifies \textit{Faithfulness} by measuring the impact of significant perturbations on the predictive function. It is computed as the MSE between the attributions multiplied by a perturbation value and the difference between the predictive function evaluated on the original input and the perturbed input. 

\begin{equation}
Inf(g,f,x) = {\mathbb{E}_{I \sim {\mu}_{I}} \Big[(I^{T} g(f,x) - (f(x) - f(x-I)))^{2}\Big]}. 
\end{equation}

Here, $I$ represents a significant perturbation around $x$. The choice of perturbation should align with the task at hand. A common choice for perturbation is Gaussian-centered noise with a specific std. We adopt this option, with an std being the average distance between dataset's points \cite{bhatt_evaluating_2020}.

$\ $ \newline
\noindent
\textbf{Sensitivity} \cite{yeh2019fidelity} measures the impact of small perturbations on the predictive function. It is computed as the gradient of the explainability function with respect to the input: 
$\nabla_{x} g(f(x))]_{j} = \lim_{\epsilon \to 0} \frac{g(f(x+\epsilon e_j)) - g(f(x))}{\epsilon}$. Here, $e_{j}$ is the basis vector of coordinate $j$. However, Sensitivity is often derived as the Max-Sensitivity within a sphere around the input of radius $r$: 

\begin{equation}
SENS_{MAX}(g,f,x,r) = \underset{\lVert y-x \rVert \le r}{max} \lVert g(f,x) - g(f,y) \rVert.
\end{equation}

Max-Sensitivity proves to be a more robust metric compared to local Lipschitz continuity measure, as the later can be unbounded for NNs. In our experiments, we sample perturbed inputs within a sphere centered around the original sample, with a radius equal to that chosen for Infidelity; the average distance between dataset's points.

\end{document}